\documentclass[10pt]{article}
\usepackage[usenames,dvipsnames]{xcolor}
\usepackage[hidelinks,colorlinks,citecolor=Fuchsia,urlcolor=blue,linkcolor=Cerulean]{hyperref}

\usepackage{times}
\usepackage{covington}
\usepackage{enumerate}
\usepackage{mdwlist}
\usepackage{natbib}

\begin{document}
\title{TimeML-strict: clarifying temporal annotation}
\author{Leon Derczynski$^{\heartsuit}$, Hector Llorens$^{\clubsuit\diamondsuit}$, Naushad UzZaman$^{\spadesuit\diamondsuit}$ \\
\small $^\heartsuit$: \texttt{leon@dcs.shef.ac.uk} University of Sheffield, UK\\
\small $^\clubsuit$: \texttt{hllorens@dlsi.ua.es} University of Alicante, Spain\\
\small $^\spadesuit$: \texttt{naushad@cs.rochester.edu} University of Rochester, USA\\
\small $^\diamondsuit$: \texttt{www.nuance.com} ~ Nuance Communications\\
}
\date{}

\maketitle

\begin{abstract}
TimeML is an XML-based schema for annotating temporal information over discourse.
The standard has been used to annotate a variety of resources and is followed by a number of tools, the creation of which constitute hundreds of thousands of man-hours of research work.
However, the current state of resources is such that many are not valid, or do not produce valid output, or contain ambiguous or custom additions and removals.
Difficulties arising from these variances were highlighted in the TempEval-3 exercise, which included its own extra stipulations over conventional TimeML as a response.

To unify the state of current resources, and to make progress toward easy adoption of its current incarnation ISO-TimeML, this paper introduces TimeML-strict: a valid, unambiguous, and easy-to-process subset of TimeML.
We also introduce three resources -- a schema for TimeML-strict; a validator tool for TimeML-strict, so that one may ensure documents are in the correct form; and a repair tool that corrects common invalidating errors and adds disambiguating markup in order to convert documents from the laxer TimeML standard to TimeML-strict.
\end{abstract}

\section{Introduction}
TimeML~\citep{pustejovsky2005specification} is an annotation scheme for the challenging task of annotation of temporal information over natural language text.
Time as expressed in language is complex and often ambiguous, and determining how to annotate it for e.g. computational processing is accordingly difficult~\citep{jaszczolt2009representing}.

Almost a decade on from its release, the TimeML schema has been adopted by hundreds of projects worldwide, and has developed into an ISO standard~\citep{PUSTEJOVSKY10.55}.
It is a comprehensive, expressive annotation markup language for temporal annotation, having been applied to significant amounts of resources and provided a framework for notably furthering research in temporal information extraction and understanding of temporal semantics.
Indeed, machine-readable temporal annotation has found applications in a wide variety of domains, including: legal~\citep{howald2011transformation,ramakrishna2011novel}, linguistic theory~\citep{derczynski-gaizauskas:2013:IWCS2013}, question answering~\citep{saquete2009enhancing,UzZamanQAEvaluation}, social media and data management~\citep{DBLP:conf/edbt/Derczynski0J13}, human interface design~\citep{uzzaman2011multimodal}, sports coverage~\citep{borg2007time}, transport accident analysis~\citep{johansson2005automatic}, working with deaf children~\citep{arfe2009evaluations}, and especially the clinical~\citep{jung2011building,sun2013evaluating}.

Over time, use cases have been discovered where it is beneficial to make some voluntary constraints regarding how TimeML is used.
Certain informal agreements were entered into by researchers who wished to exchange data reliably.
Recently, these constraints have been formally applied in the 3rd international temporal evaluation exercise, TempEval-3~\citep{DBLP:journals/corr/abs-1206-5333}, where the corpus\footnote{Available in the ACL Data and Code Repository, reference \href{http://aclweb.org/aclwiki/index.php?title=TempEval-3_Platinum_TimeML_annotations_(Repository)}{ADCR2013T001}} followed such agreements.
This paper details and crystallises these constraints into a voluntary, formal standard describing a subset of TimeML, called TimeML-strict.

As well as serving as a canonical reference of the standard used in TempEval-3, this paper also introduces arguments for TimeML-strict, and a range of supporting tools.
We describe a schema for TimeML-strict; a validation tool for checking whether a TimeML document is conformant; and a repair tool, for tightening up existing data such that is it compliant.
Our hope is that compliant data becomes easier to automatically process, thus widening the potential user base of TimeML as well as easing interoperability~\citep{lee2010towards}.
The resources presented should also aid automatic conversion of legacy temporal annotation to ISO-TimeML, increasing the useful lifetime of community linguistic resources that represent large amounts of effort and investment in semantic annotation.

The paper is structured as follows.
Section~\ref{sec:problems} discusses common difficulties encountered by end-users of TimeML.
Section~\ref{sec:changes} describes the changes in TimeML-strict.
Section~\ref{sec:resources} describes supplementary resources  and Section~\ref{sec:not-covered} makes explicit some things that TimeML-strict does not address.
We conclude in Section~\ref{sec:conclusion}.

\section{Problems with the current state of resources}\label{sec:problems}

What follows is a selection of common technical issues encountered when processing TimeML, followed by a requirements specification for a standard clarification.

\subsection{Validity}

As an XML format, TimeML documents can be validated by means of an XML ``document type description", which is part of the TimeML standard.
However, not all resources currently distributed as TimeML are valid according to their XML document type description (DTD).
This gives many problems when loading data as XML, via e.g. the document object model (DOM).
The inability to load TimeML documents as XML directly removes many of the advantages of choosing XML for the standard -- such as ease of use with text processing frameworks~\citep{ogren2008cleartk,cunningham2013getting} -- and costs many man-days a year per researcher working with invalid data.
Errors range from mis-typed element named to wrongly-encoded characters (from e.g. other alphabets) to SGML-valid but structurally inconsistent documents (e.g. where two TIMEX3 annotations have the same ``unique" ID).
Many different tools have made parallel efforts to overcome these difficulties.
However, all these efforts would not be required if content creators published valid data in the first place.

\subsection{Timestamps}
It is sometimes unclear what the document creation time is.
TimeML's TIMEX3 functionInDocument attribute is there to label whether a timex is publication date, creation date and so on -- an expressive and useful part of the schema, permitting capture of multiple important dates which often act in two rules, both as timexes in discourse and as document meta-information.
Often, only one of these special-function dates is specified (creation time or publication date), though sometimes none is, and sometimes more than one is.
However, there is no way of defining which of these dates should be used as the default reference time.
Having such a definition is critical to timex normalisation~\citep{LLORENS12.128} as well as to accurate replication of results.

\subsection{What to annotate}
TimeML is often used to annotate newswire documents (e.g., in the biggest TimeML corpus, TimeBank~\citep{TimeBank}; in the TempEval-3 corpus; and also in the AQUAINT TimeML corpus).
A lot of these are taken verbatim from the source, and include a preamble of metadata that is essentially gibberish -- certainly not natural language (Section~\ref{sec:text} below contains an example).
This non-linguistic content often gets in the way of working with various NLP tools: it is often unclear how this preamble should be treated.
Does one count it as a single sentence?
Should headlines and editorial comments within it be annotated?
How about numerically encoded dates that occur as fragments?

\subsection{Inconsistencies}
Many documents are produced that appear consistent but are difficult to process.
These may include, for example, non-standard id labels; EVENT elements should always have an eid attribute that takes the form of {\tt eXX} where XX is a positive integer, though some tools treat the field as freeform text, or omit the {\tt e}.
In other cases, edits to a document may create partial information (through both deletion and insertion of annotations).
For example, after deleting an event instance, an ALINK may still use that event instance ID as one of its arguments.
All these phenomena create speed bumps working with TimeML, but can be readily checked for. 

\subsection{Requirements}
The majority of problems above stem from the organic way in which tools and resources have sprung up over the past years, all using the framework of TimeML.
We hope to provide a common ground -- and means of reaching it -- that is both TimeML-compatible and also very easy to work with.
One goal of this standard refinement is to ease the process of programming with TimeML.
Therefore, it should be easy to implement systems that accept TimeML-strict.
For acceptance of TimeML-strict to be a sufficient programming requirement for working with TimeML, it is important that legacy data and annotations produced by legacy tools are compatible with systems that expect TimeML-strict.
Finally, one must not constrain the expressiveness of TimeML: rather, the goal it to carefully preserve this expressiveness, and maximise access to TimeML-annotated resources.

\section{Changes new in TimeML-strict}\label{sec:changes}

This section details our proposed annotations, as an addendum to the TimeML standard v1.2.
Parts of ISO-TimeML are accepted either, though the standard is not as well-used as TimeML, and partially in a state of flux.
TimeML-strict does include some features of ISO-TimeML: most notably,

\begin{itemize}
\item instantiating events by including {\tt eiid} and other event instance attributes within an EVENT label, thus eliminating the need for MAKEINSTANCES for the majority case where events are instantiated once;
\item support event verb form and predicate attributes ({\tt vForm} and {\tt pred}).
\end{itemize}

\subsection{The {\tt DCT} element}

This is introduced to resolve potential ambiguity regarding the document's creation time, that should be used as a default anchor for timexes within the document.
There must be exactly one DCT per document.
This element should enclose a single TIMEX3 element, with no other intervening nodes (including text nodes) -- see Example~\ref{ex:dct}.

\begin{example}
\label{ex:dct}
{\tt <DCT><TIMEX3 functionInDocument="CREATION\_TIME" temporalFunction="false" tid="t0" type="DATE" value="2013-03-22">March 22, 2013</TIMEX3></DCT> }
\end{example}

In the case of documents where DCT is not known, not given, or otherwise unclear, give an underspecified self-closing day-level timex annotation (Example~\ref{ex:unknown-dct}).

\begin{example}
\label{ex:unknown-dct}
\texttt{<DCT><TIMEX3 tid="t0" value="XXXX-XX-XX" /></DCT>}
\end{example}

\subsection{The {\tt TEXT} element}\label{sec:text}

This is used to specify exactly the bounds of linguistic content in the document.
The goal is to be clear which content should be considered for annotation, and allow exclusion of non-linguistic document content.
Example~\ref{ex:text} shows how this element can be used to exclude newswire preamble.

\begin{example}
\footnotesize
\begin{verbatim}
<TimeML>

 AP900815-0044 
AP-NR-08-15-90 1337EDT
u i PM-GulfRdp 8thLd-Writethru   08-15 1334
PM-Gulf Rdp, 8th Ld-Writethru,a0605,1368
Saddam Seeks End To War With Iran; Bush To Urge Jordan To Close
Port
Eds: SUBS 28th graf pvs, Crown Prince... to CORRECT spelling of
Hassan; pick up 29th graf pvs, `A CBS...'
LaserPhotos WX6,7,XSAV1,NY5,10,TOK1,XAAFB1,AMM1, LaserColor XAAFB1
By CHRISTOPHER BURNS
Associated Press Writer

<TEXT>
   Iraq's Saddam Hussein, <EVENT eid="e5" 
class="STATE">facing</EVENT> U.S. and Arab troops at the Saudi

...

</TEXT>

</TimeML>
\end{verbatim}
\label{ex:text}
\end{example}

Each document must have exactly one {\tt TEXT} element.
No particular XML hierarchical relation between the {\tt TEXT} and {\tt DCT} elements is required; {\tt DCT} may be before, after, within, or even contain the {\tt TEXT} element (i.e. parent, child and sibling are all fine).

\subsection{Schema validation requirement}
All TimeML documents should include a reference to the TimeML DTD, in order to assist with their validation.
This requirement has not been sufficient to ensure clean, legible XML.
TimeML-strict documents \emph{must} be valid.

As well as TimeML DTD validation, TimeML-strict also requires that documents validate according to a more rigourous XML schema: the TimeML-strict XSD.
This is the core strict requirement, intended to aid processing of TimeML by other tools.
Schema compliance make transforming TimeML-strict to other formats (such as those following the ISO Language Annotation Framework, including ISO-TimeML) easy, with one-off transformation descriptions via e.g. XSLT.
Following the standard enables easy conversion to ISO-TimeML whenever updates are released to the community (through formal XML translations, instead of text processing or intricate SAX event handlers and so on). 

A TimeML validation tool is made available in order to assist meeting this requirement, and the corresponding XML schema file (see Section~\ref{sec:validator}).

\subsection{No phantom element IDs}
XML schemas can enforce checks for missing references and elements.
For example, a TLINK may reference a nonexistent event instance; elements may have malformed identifiers.
TimeML-strict requires that every reference to an element be an identifier giving reference to a findable element.

\subsection{Semantics of DURING}

In his 1983 account, Allen determined a small set of distinct possible relations between two temporal orderings~\citep{allen1983maintaining}.
TimeML uses this full-interval temporal relation set for TLINKs.
For the most part, the relations in Allen's paper and in TimeML are simple to match.
TimeML introduces the IDENTITY relation, to distinguish between events (or times) that happen at the same time, and those that are also the same thing.
However, Allen's {\tt overlap} and {\tt overlap-inverse} relations do not seem to be accounted for in TimeML.
Similarly, the TimeML DURING / DURING\_INV relations are not defined strongly in the annotation guidelines or v1.2 specification.

The apparent absence of Allen's {\tt overlap} relations leaves a hole in TimeML's expressiveness.
See Example~\ref{ex:overlap} -- there is no other temporal relation that would be appropriate between e1 and t1.
The winter starts within 2012, but continues beyond its termination.
The relation is not inclusion or simultaneity, and both BEGINS and ENDS (and their inverses) require a shared interval endpoint, which is also not the case here.

\begin{example}
\footnotesize
\begin{verbatim}
The <EVENT eid="e1" eiid="ei1" class="OCCURRENCE">winter</EVENT> that
started in <TIMEX3 tid="t1" type="DATE" value="2012">2012</TIMEX3>
was one of the coldest on record.
\end{verbatim}
\label{ex:overlap}
\end{example}

Further, the DURING relation seems to duplicate functionality or either INCLUDES or SIMULTANEOUS, depending on subjective interpretation.
One popular definition is that it should be used when an event occurs during a timex.
However, this functionality is already explicitly provided by SIMULTANEOUS / IS\_INCLUDED.
Choosing one of those gives a more precise relation.
It is unclear why one would want to omit Allen's overlap functionality but include an underspecified superlabel for just one specific circumstance.
Also, the idea that some relation labels can only apply to particular types of intervals seems unintuitive, and is a departure from the general theme in TimeML of abstracting events and timexes to intervals.

Finally, it is difficult to introduce a new {\tt relType} value to TimeML for these relations; doing so breaks TimeML compatibility, and signifies a departure from TimeML, which is the opposite of our goal.

In TimeML-strict, this apparent mismatch is interpreted as an oversight, and the two explicitly mapped as follows.

\begin{enumerate}
\item TimeML DURING is equivalent to Allen's ``{\tt overlap inverse (oi)}": A DURING B is read as \emph{``A starts during B and persists beyond the end of B"} (e.g. overlap where A starts within B);
\item TimeML DURING\_INV is equivalent to Allen's ``{\tt overlap (o)}": A DURING\_INV B is read as \emph{``During A, B starts and persists beyond the end of A"} (e.g. overlap where B starts within A).
\end{enumerate}

Under TimeML-strict, for the intervals in Example~\ref{ex:overlap}, one might annotate:

\footnotesize
\begin{verbatim}
<TLINK lid="l1" eventInstanceID="ei1" relType="DURING"
 relatedToTime="t1" />
\end{verbatim}
\normalsize

This change is perhaps in conflict with some prior interpretations of TimeML, but it is the only way in which TimeML's relation types can be made to cover the full set of interval relation configurations and removes two arguably redundant links, while staying valid.

As this change risks taking a departure from some conventions, one should pay regard to impact on existing resources.
Brief examination of TimeBank 1.2 suggested annotator ambiguity between all of the INCLUDES, SIMULTANEOUS and DURING link relation types.
We do not specify how to deal with such annotation, and recommend that annotations (especially in resources) be treated as slightly fuzzy, as per the TimeML recommendation.

Existing closure tools should be readily modifiable to meet this specification change and indeed some have already been using this interpretation for a few years.

\section{Resources provided}\label{sec:resources}

With TimeML-strict, three resources are made available for working with documents and the specification.

\subsection{XML Schema Definition}

Core to TimeML-strict is a formal validation schema.
This is substantially different from TimeML 1.2's XML schema definition (XSD), building in stricter checks.
There are two key scenarios for this schema's use: to help TimeML producers be sure that they have generated shareable, legible TimeML; and to enable those trying to read TimeML to be aware of the expected breadth of expression and level of data consistency.

This schema includes TIMEX3 support.
There is no single reference point for the TIMEX3 standard, but rather TimeML builds on earlier TIMEX standards.
The TimeML-strict schema incorporates the TIMEX2 spec from~\citet{ferro2005tides} and the appropriate adaptations in TimeML 1.2.
The schema is included with the TimeML validator.

\subsection{TimeML validator}\label{sec:validator}

A Java validation tool that verifies whether or not documents are acceptable as TimeML-strict.
For documents that are not valid, information is given to allow content creators to find the source of the problems.
This tool is available via github.\footnote{See \href{https://github.com/hllorens/TimeML-validator}{https://github.com/hllorens/TimeML-validator}}

\subsection{Automatic migration and repair of TimeML}

To ease adoption of TimeML-strict, and to convert potentially invalid data into a consistent, valid format, a TimeML repair tool is made available.
This converts and fixes common mistakes in older TimeML, and enables older resources to be processed by newer tools. 

The DCT and TEXT elements are automatically added by the tool, if possible, which also attempts to rectify several common mistakes such as invalidly-structured ID strings or references to missing entities.
As TimeBank v1.2 isn't valid under TimeML-strict, this tool is absolutely critical.
The tool has also been tested with a few common TimeML-generating systems and can rectify their output, repairing both invalid TimeML and also adding requisite information to make the output TimeML-strict compliant.
It is accessible via open-source repository.\footnote{See \href{https://bitbucket.org/leondz/timeml-repair}{https://bitbucket.org/leondz/timeml-repair}}

\section{Beyond the scope of TimeML-strict}\label{sec:not-covered}

This paper has so far considered problems encountered ``in the wild" and proposed fixes for them.
There are also those problems that are deliberately \emph{not} addressed by TimeML-strict.
This section introduces a few of these phenomena with explanations of why TimeML-strict does not constraint against them.

The extent of timex and event elements should be just a single word, according to the TimeML English annotation guidelines.
However, there are many cases where this is not possible, especially with timexes, where qualifying words are critical (e.g. \emph{last} in \emph{``last year"}.
Also, some resources migrated from older standards include slightly longer words~\citep{DERCZYNSKI12.451.L12-1237}.
Making this invalid in TimeML-strict would be an unreasonable constraint and would reduce the expressiveness of TimeML, while offering little benefit to the technical process (the annotated text is e.g. a single DOM node regardless of word count).

TimeML-strict also does not tackle temporal consistency.
The transitive nature of many interval relation types means that it is possible to create an annotation that is inconsistent (Example~\ref{ex:inconsistent}).

\begin{example}
\footnotesize
\begin{verbatim}
<TLINK lid="l1" eventInstanceID="ei1" relType="BEFORE"
  relatedToEventInstance="ei2" />
<TLINK lid="l2" eventInstanceID="ei1" relType="INCLUDES"
  relatedToEventInstance="ei2" />
\end{verbatim}
\label{ex:inconsistent}
\end{example}

Although this presents difficulties when performing temporal closure or attempting to use inference-based learning for relation labelling, such annotations remain valid.
TimeML should be capable of annotating any document.
In the case of newswire, one should not make the assumption that the linguistic utterances of journalists are temporally consistent in the first place.

It is also possible to create somewhat ``orphaned" annotations, e.g. uninstantiated events.
There is no requirement to instantiate events, and inclusion of any event attribute other than its class is optional. 
Indeed, TimeML's possible values for part of speech, tense, aspect and so on may be viewed as recommendations rather than declarations of the best way to annotate this values for temporal information processing, reducing how critical this information is.

Some of these kinds of meta-consistency is explicitly checked for by the CAVaT tool~\cite{derczynski2010cavat}, which includes a selection of modules for validating TimeML.
However, they are all permissible under TimeML-strict.

\section{Conclusion}\label{sec:conclusion}

This paper has detailed a refinement of the TimeML standard, hoping to bring together the many diverse outputs of the successful ongoing TimeML project and make life easier for those working with TimeML data.
Along with the abstract parts of the refinement, three concrete tools are presented: a schema; a validation tool; and a repair tool.
This is an iterative, voluntary step, which has the added benefit of preparing for rapid and seamless transition between TimeML and ISO-TimeML.
TimeML-strict offers a common base for simpler computing with temporal annotations, allowing interested researchers to get on with experimentation and discovery.

\section*{Acknowledgements}

The authors would like to thank James Pustejovsky, Marc Verhagen and Rob Gaizauskas for helpful discussions over the past years.
We also thank the TempEval-3 participants~\citep{uzzaman2013semeval} for their detailed feedback and for being superlative test subjects.
Part of this work was supported by CHIST-ERA EPSRC grant No. EP/K017896/1 uComp,\footnote{See \href{http://www.ucomp.eu/}{http://www.ucomp.eu/}} and also by Aarhus University, Denmark who kindly provided facilities.

\bibliographystyle{te}
\bibliography{tmlstrict}

\end{document}